\documentclass[letterpaper, 10 pt, conference]{ieeeconf}  

\IEEEoverridecommandlockouts                              

\usepackage{graphics} 
\usepackage{epsfig} 
\usepackage{times} 
\usepackage{amsmath} 
\usepackage{amssymb}  
\usepackage{cite}
\usepackage{amsmath,amssymb,amsfonts}
\usepackage{algorithmic}
\usepackage{graphicx}
\usepackage{textcomp}
\usepackage{xcolor}
\usepackage{lipsum}
\usepackage{array}
\usepackage{booktabs}
\usepackage{tabularx}
\usepackage{multicol}
\usepackage{multirow}
\usepackage{xcolor}
\usepackage{amssymb}
\usepackage{amsmath}
\usepackage{stfloats}
\usepackage{mathdots}
\usepackage{graphicx} 
\usepackage{caption}
\usepackage{subfigure}
\usepackage{algorithm}
\usepackage{bm}
\usepackage{listings}
\usepackage{hyperref}

\title{\LARGE \bf
Robot Shape and Location Retention in Video Generation Using Diffusion Models
}

\author{Peng Wang, Zhihao Guo, Abdul Latheef Sait, Minh Huy Pham
\thanks{*This work was supported by the Department of Computing and Mathematics, Manchester Metropolitan University.}
\thanks{Peng Wang ({\tt\small p.wang@mmu.ac.uk}), Zhihao Guo, Minh Hui Pham are with the Department of Computing and Mathematics, 
        Manchester Metropolitan University, M15 6BH Manchester, UK.
        Abdul Latheef Sait is with JD Sports Fashion PLC, and the work was done while at MMU.
        }
}

\begin{document}

\maketitle
\thispagestyle{empty}
\pagestyle{empty}

\begin{abstract}

Diffusion models have marked a significant milestone in the enhancement of image and video generation technologies. However, generating videos that precisely retain the shape and location of moving objects such as robots remains a challenge. This paper presents diffusion models specifically tailored to generate videos that accurately maintain the shape and location of mobile robots. This development offers substantial benefits to those working on detecting dangerous interactions between humans and robots by facilitating the creation of training data for collision detection models, circumventing the need for collecting data from the real world, which often involves legal and ethical issues. Our models incorporate techniques such as embedding accessible robot pose information and applying semantic mask regulation within the ConvNext backbone network. These techniques are designed to refine intermediate outputs, therefore improving the retention performance of shape and location. Through extensive experimentation, our models have demonstrated notable improvements in maintaining the shape and location of different robots, as well as enhancing overall video generation quality, compared to the benchmark diffusion model. Codes will be opensourced at \href{https://github.com/PengPaulWang/diffusion-robots}{Github}.

\end{abstract}

\section{INTRODUCTION}

Diffusion models have achieved remarkable advancements in recent years, and have achieved better or on-the-par performance with generative adversarial networks in image and video generation~\cite{cao2022survey,croitoru2023diffusion}. Compared to image generation, video generation remains a challenge in terms of model complexity, dependence on data and computational resources, consistency of generated videos, generation efficiency, and shape and location retention of dynamic objects in generated videos~\cite{hoppe2022diffusion, ho2022video}. Despite all the challenges, the potential of diffusion models to generate dynamic and appealing content has driven the research and application forward, and they have been applied in generating high-quality videos~\cite{ho2022video}, carrying out video prediction and infilling~\cite{hoppe2022diffusion}, control movements in the generated video~\cite{chen2023motion}, and directly process and manipulate a real input video~\cite{nikankin2022sinfusion}. Another promising application of diffusion models is that they can be used to generate data for dangerous interaction detection in cases such as human-robot collaboration, where collecting real data for model training faces legal and ethical challenges.

The foundational technology behind many of the applications mentioned is the Denoising Diffusion Probabilistic Model (DDPM), which is trained to understand Gaussian noise patterns added to input images throughout the training process. Once sufficiently trained, the DDPM can start with noisy images or images that consist purely of Gaussian noise and, through iterative denoising, produce outputs that adhere to a specific empirical distribution~\cite{sohl2015deep, ho2020denoising,ho2022video, wang2022semantic}.

\begin{figure}[t]
    \centering \includegraphics[width=1.0\linewidth]{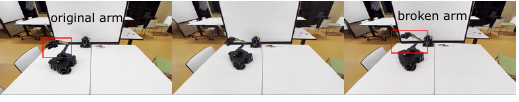}
    \caption{The original and generated frames of a robot. \textit{Left:} the original frame; \textit{Middle:} the frame generated by a proposed model; \textit{Right}: the frame generated by the benchmark model. The robot arm is broken in the frame generated by the benchmark model.}
    \label{fig:framesgrid}
\end{figure}
The evaluation of diffusion models' performance often relies on metrics like the Peak Signal-to-Noise Ratio (PSNR), which measures the overall quality of frames or videos by computing pixel-to-pixel differences between the generated frames and the reference frames if any. However, relying solely on PSNR may overlook structural information loss, such as local distortions of the shape of objects of interest, providing a misleadingly positive assessment of overall performance. For instance, Figure~\ref{fig:framesgrid} shows one original frame (\textit{left}) with a robot, and two frames generated by diffusion models (\textit{middle} and \textit{right}). We can see that the generated frame on the right has a broken arm (lost retention of the shape), while the generated frame in the middle maintains the shape of the arm. Despite the failed arm shape retention, the two generated frames have similar PSNR values as the distorted arm does not contribute enough to make a distinctive difference in PSNR values. This oversight is particularly critical in scenarios where an object's shape and location are crucial in generated frames. For instance, in human-robot collaborative tasks, there is the need to forecast potential collisions between humans and (dynamic) robots, and the collection of such data for collision model training in real life often faces ethical and legal challenges. Therefore, using diffusion models to generate data with shape and location retention becomes a promising solution. In light of these observations, the Structural Similarity Index (SSIM) emerges as an alternative metric for evaluating diffusion models. Unlike PSNR, SSIM is adept at capturing structural similarities and differences, making it a more reliable indicator of a model's ability to preserve object shapes and locations. 

This paper aims at developing diffusion models that can generate frames where the shape and location of objects of interest can be retained. Particularly, we are interested in generating videos that contain moving robots, whose shape and location retention are vital in the generated frames. As mentioned earlier, this will for example help to generate data for human-robot collision detection tasks and bypass legal and ethical challenges. Two types of robots are used in different scenarios, i.e., a Waffle Pi mobile robot with a gripper mounted on top and a collaborative robot, a.k.a., cobot. The proposed diffusion models take the ConvNext~\cite{liu2022convnet} as the backbone network, to accelerate the training and sampling efficiency~\cite{nikankin2022sinfusion}. To retain the shape and location of the robots, we have embedded the robot pose information such as location, orientation, and velocities into ConvNext blocks and used semantic masks (either the masks of the robots or the masks of the robots and the backgrounds) to regulate the intermediate outputs of ConvNext blocks. Various experiments have been conducted to investigate how pose embedding and mask regulation affect the performance of the models in shape and location retention.

The contributions of this work include 1) the development of diffusion models capable of preserving the shape and location of robots within generated frames. This advancement shows promise for generating data to train models aimed at detecting collisions between humans and robots in the future. 2) the introduction of a novel Spatially-Adaptive Normalization (SPADE) module for integrating semantic masks, and the implementation of an embedding procedure that incorporates robot pose information from controllers like the Robot Operating System (ROS) into the backbone network, which strikes a balance between the quality of generation and the preservation of shape and location information. 3) Introduction of a refined Intersection over Union (IoU) metric and the Hu moments match for evaluating the retention of location and shape.

The remainder of the paper is organised as follows: Section II presents some related works, Section III elaborates on the approach, Section IV covers experiments, discussions, and an ablation study, and finally, Section V concludes the paper.

\section{RELATED WORKS}\label{sec:relatedwork}

Most video generation models based on DDPMs share the same underlying core backbone UNet~\cite{ronneberger2015u,ho2022video} for the denoising process. They substantially differentiate from each other in terms of conditions for generating new frames. There are mainly three types of conditions, i.e., (i) embedded context information: For example, Yang et al.~\cite{yang2022diffusion} propose residual video diffusion, which utilises a context vector generated by a convolutional recurrent neural network as conditions to generate the next frame.
embedded context information, e.g., Yang et al~\cite{yang2022diffusion} propose   (ii) semantic masks, e.g., Wang et al~\cite{wang2022semantic} propose the semantic diffusion model where semantic masks are employed to condition new frames, resulting in improved quality in generating small objects; and (iii) video frames as conditions, e.g., Vikram et al~\cite{voleti2022mcvd} propose masked conditional video diffusion, which involves masking frames from the past or future. The model is trained on unmasked frames and generates the masked frames based on the chosen masking strategy.

Yaniv et al.~\cite{nikankin2022sinfusion} have recently introduced SinFusion, a video generation diffusion model utilizing ConvNext~\cite{liu2022convnet} as the backbone. This model can produce images or videos based on a single input image or video. The novel architecture proves particularly advantageous in training DDPMs on a single image or its large crops, circumventing the `overfitting' issues associated with UNet. This is achieved by restricting the receptive field of UNet to non-global areas and reducing computational time compared to standard DDPMs~\cite{ho2020denoising}. Such improvements are especially beneficial for real-life applications like human-robot collaboration~\cite{wang2024deep}.

While SinFusion has demonstrated comparable or even superior results compared to other video generation models, the authors note potential drawbacks, such as the possibility of breaking dynamic objects in the generated results. An example of this issue is illustrated in Figure~\ref{fig:framesgrid} (\textit{Right}).

Inspired by the simple structure and efficiency of SinFusion, and the embedding techniques from other works~\cite{wang2022semantic} in improving the performance of diffusion models, we have adopted ConvNext as our backbone and introduced robot pose embedding and semantic mask regulation to help retain the shape and location of dynamic object (robots) generation, a step towards applying diffusion models to generate data for dangerous human-robot interaction detection model training.

\section{DIFFUSION MODELS}

The theory and fundamental principles of diffusion models were introduced by Sohl-Dickstein et al.\cite{sohl2015deep} and further elaborated upon in subsequent studies like those by Ho et al.\cite{ho2020denoising,ho2022video}, as well as other works such as that by Hoppe et al.\cite{hoppe2022diffusion}. In essence, diffusion models utilise a deep neural network $\mathcal{M}$, such as UNet\cite{ronneberger2015u}, as their backbone network. This network is trained on noisy data, such as images and video frames, to enable the trained model to accurately identify and model the noise present in the input data.

The training of diffusion models comprises two primary stages: the forward diffusion process (forward process) and the reverse diffusion process (reverse process). In the forward process, data such as images and videos serve as inputs, and the structure of the data distribution is disrupted by introducing noise. This facilitates the training of model $\mathcal{M}$ to recognize and model the noise imposed on the data. The reverse diffusion process, known as the reverse process, aims to reconstruct the data structure from noisy data or the noise itself. In this paper, we will first review these two stages of diffusion models in the context of video generation, followed by our proposed works.

\subsection{The Forward Diffusion Process}

In the context of image/video generation, given an input frame $\mathbf{x}_0$ sampled from a distribution $q(\mathbf{x}_0)$, one can iteratively add Gaussian noise $\boldsymbol{\Sigma}_t \sim \mathcal{N}(\boldsymbol{\Sigma}_t;\mathbf{0},\mathbf{I})$, $t=1,\cdots, T$ to $\mathbf{x}_0$ for $T$ steps. This process generates a sequence of noisy samples $\{\mathbf{x}_1, \cdots, \mathbf{x}_T\}$. The variance of the noise added at each step can be controlled using a variance scheduler $\{\beta_t \in (0,1)\}_{t=1}^T$. The forward diffusion process is normally formulated as a Markov chain:
\begin{equation}\label{eq:forward}
    q(\mathbf{x}_{1:T}|\mathbf{x}_0):=\prod_{t=1}^T q(\mathbf{x}_t|\mathbf{x}_{t-1}),
\end{equation}
\noindent
where
\begin{equation}\label{eq:forwardDiffusion}
    q(\mathbf{x}_t \vert \mathbf{x}_{t-1}) := \mathcal{N}\big(\mathbf{x}_t; \sqrt{1 - \beta_t} \mathbf{x}_{t-1}, \beta_t\mathbf{I}\big),
\end{equation}
\noindent
which indicates the dependency of $\mathbf{x}_t$ on $\mathbf{x}_{t-1}$. This also implies that to get a noisy sample at $\mathbf{x}_t$, one needs to add noises from $\mathbf{x}_0$ up to $\mathbf{x}_{t-1}$ step by step, which could be time and computational resources demanding. Fortunately, this can be simplified as shown in~\cite{ho2020denoising}, i.e., the forward process admits sampling $\mathbf{x}_t$ at an arbitrary timestep $t$ in closed form. This is achieved by letting $\alpha_t = 1 - \beta_t$ and $\bar{\alpha}_t = \prod_{i=1}^t \alpha_i$, one then gets 

\begin{equation}\label{eq:onestep}
    q(\mathbf{x}_t \vert \mathbf{x}_0) := \mathcal{N}\big(\mathbf{x}_t; \sqrt{\bar{\alpha}_t} \mathbf{x}_0, (1 - \bar{\alpha}_t)\mathbf{I}\big),
\end{equation}
\noindent
which indicates that $\mathbf{x}_t$ can be sampled from $\mathbf{x}_0$ in one step as in 
\begin{equation}\label{eq:reparameter}
\mathbf{x}_t 
= \sqrt{\bar{\alpha}_t}\mathbf{x}_0 + \sqrt{1 - \bar{\alpha}_t}\boldsymbol{\Sigma}, 
\end{equation}
\noindent
where $\boldsymbol{\Sigma}\sim \mathcal{N}(\mathbf{0},\mathbf{I})$ is the noise used to generate the noisy frame $\mathbf{x}_t$.

\subsection{The Reverse Diffusion Process}

The reverse diffusion process involves starting with a Gaussian noise $\mathbf{x}_T\sim \mathcal{N}(\mathbf{0},\mathbf{I})$ and then reversing the transition outlined in Equation (\ref{eq:forward}). This reversal allows for sampling from the posterior of the forward process $q(\mathbf{x}_{t-1}|\mathbf{x}_t)$, with $t = T, \cdots, 1$, in order to recover $\mathbf{x}_0$ (it's worth noting that the process can terminate at any intermediate stage). However, reversing Equation (\ref{eq:forward}) presents a challenge, and it is typically approximated using a trainable Markov chain depicted in Equation (\ref{eq:reverse}), which begins with a Gaussian noise $p(\mathbf{x}_T)=\mathcal{N}(\mathbf{x}_T;\mathbf{0},\mathbf{I})$:
\begin{equation}\label{eq:reverse}
    p_\theta(\mathbf{x}_{0:T}):=p(\mathbf{x}_T)\prod_{t=1}^T p_\theta(\mathbf{x}_{t-1}|\mathbf{x}_t),
\end{equation}
\noindent
where
\begin{equation}
    p_\theta(\mathbf{x}_{t-1}|\mathbf{x}_t):=\mathcal{N}\big(\mathbf{x}_{t-1};\boldsymbol{\mu}_\theta(\mathbf{x}_t, t), \boldsymbol{\Sigma}_\theta(\mathbf{x}_t,t)\big).
\end{equation}

One can see that if $p_\theta(\mathbf{x}_{0:T})$ can be learned by $\mathcal{M}$, then the reverse process simplifies to 
\begin{equation}\label{eq:simplifiedreverse}
    p_\theta(\mathbf{x}_0):=\int p_\theta(\mathbf{x}_{0:T})d\mathbf{x}_{1:T},
\end{equation}
where $\mathbf{x}_{1:T}$ are latent variables of the same dimensions with $\mathbf{x}_0$. The approximation of $q(\mathbf{x}_{1:T}|\mathbf{x}_0)$ using $p_\theta(\mathbf{x}_{0:T})$ is achieved by optimising the variational bound on negative log-likelihood between them~\cite{ho2020denoising}:
\begin{equation}
    \mathbb{E}[-\log p_\theta(\mathbf{x}_0)]\leq \mathbb{E}_q\Bigg[-\log \frac{p_\theta(\mathbf{x}_{0:T})}{q(\mathbf{x}_{1:T}|\mathbf{x}_0)}\Bigg]:=L,
\end{equation}
\noindent
which can be rewritten into Equation (\ref{eq:Lexpand}) according to \cite{sohl2015deep}:
\begin{equation}\label{eq:Lexpand}
\begin{aligned}
L := 
& \mathbb{E}_q \Big[\underbrace{D_\text{KL}\big(q(\mathbf{x}_T \vert \mathbf{x}_0) \parallel p_\theta(\mathbf{x}_T)\big)}_{L_T} \\
&+ \sum_{t=2}^T \underbrace{D_\text{KL}\big(q(\mathbf{x}_{t-1} \vert \mathbf{x}_t, \mathbf{x}_0) \parallel p_\theta(\mathbf{x}_{t-1} \vert\mathbf{x}_t)\big)}_{L_{t-1}} \\
&\underbrace{- \log p_\theta(\mathbf{x}_0 \vert \mathbf{x}_1)}_{L_0} \Big],
\end{aligned}
\end{equation}
\noindent
where $D_\text{KL}$ represents the KL divergence. One can see that each term in Equation (\ref{eq:Lexpand}) is a direct measure of the similarity in terms of KL divergence between $p_\theta(\mathbf{x}_{t-1}|\mathbf{x}_t)$ and the reversed forward transitions but conditioned on $\mathbf{x}_0$, i.e., $q(\mathbf{x}_{t-1} \vert \mathbf{x}_t, \mathbf{x}_0)$. It is noteworthy that $q(\mathbf{x}_{t-1} \vert \mathbf{x}_t, \mathbf{x}_0)$ is tractable and this makes optimisation of $L$ viable, henceforth making the approximation of $q(\mathbf{x}_{1:T}|\mathbf{x}_0)$ using $p_\theta(\mathbf{x}_{0:T})$ viable.

In the context of video generation, an arbitrary noisy sample $\mathbf{x}_t$, $t = T, \cdots, 1$ sampled using Equation (\ref{eq:reparameter}) is fed to the deep neural network-based model $\mathcal{M}$, which is trained (by optimising Equation (\ref{eq:Lexpand})) to approximate the noise $\boldsymbol{\Sigma}_t$ imposed. When well trained,  $\mathcal{M}$ will be able to identify and model the noises, helping to remove the noise and restore data structures.

Inspired by advancements in image and video generation, researchers have introduced various diffusion models. These models include those that utilise semantic masks as conditions to produce high-quality images~\cite{wang2022semantic}, among others. Semantic masks offer valuable information, such as object shapes and locations, making them ideal for generative tasks that prioritise retaining shape and spatial details. Denoting conditions like masks as $\mathbf{y}$, Equation (\ref{eq:reverse}) can be reformulated as:
\begin{equation}\label{eq:reverseFactorised}
    p_\theta(\mathbf{x}_{0:T}\vert \mathbf{y}) = p(\mathbf{x}_T) \prod^T_{t=1} p_\theta(\mathbf{x}_{t-1} \vert \mathbf{x}_t, \mathbf{y}),
\end{equation}
\noindent
where
\begin{equation}\label{eq:reverseProcess}
    p_\theta(\mathbf{x}_{t-1} \vert \mathbf{x}_t, \mathbf{y}) = \mathcal{N}(\mathbf{x}_{t-1}; \boldsymbol{\mu}_\theta(\mathbf{x}_t,\mathbf{y}, t), \boldsymbol{\Sigma}_\theta(\mathbf{x}_t,\mathbf{y}, t)).
\end{equation}

Since the condition $\mathbf{y}$ applies to $p_\theta(\mathbf{x}_{t-1} \vert \mathbf{x}_t, \mathbf{y})$ for $t=T,\cdots,1$, it is straightforward to substitute these terms involve $\mathbf{y}$ into Equation (\ref{eq:Lexpand}) to derive the optimization term for conditioned diffusion models:
\begin{equation}\label{eq:LexpandConditions}
\begin{aligned}
L = 
& \mathbb{E}_q \Big[\underbrace{D_\text{KL}\big(q(\mathbf{x}_T \vert \mathbf{x}_0) \parallel p_\theta(\mathbf{x}_T)\big)}_{L_T} \\
&+ \sum_{t=2}^T \underbrace{D_\text{KL}\big(q(\mathbf{x}_{t-1} \vert \mathbf{x}_t, \mathbf{x}_0) \parallel p_\theta(\mathbf{x}_{t-1} \vert\mathbf{x}_t, \mathbf{y})\big)}_{L_{t-1}} \\
&\underbrace{- \log p_\theta(\mathbf{x}_0 \vert \mathbf{x}_1,\mathbf{y})}_{L_0} \Big].
\end{aligned}
\end{equation}
When the model is well trained, it will take in a Gaussian noise image $\mathbf{x}_T \sim \mathcal{N}(\mathbf{0},\mathbf{I})$ and `recreate' samples from it by removing the noise step by step.

\section{Shape and Location Retention Diffusion Models}

 \begin{figure*}[ht]
    \centering    \includegraphics[width=0.65\linewidth]{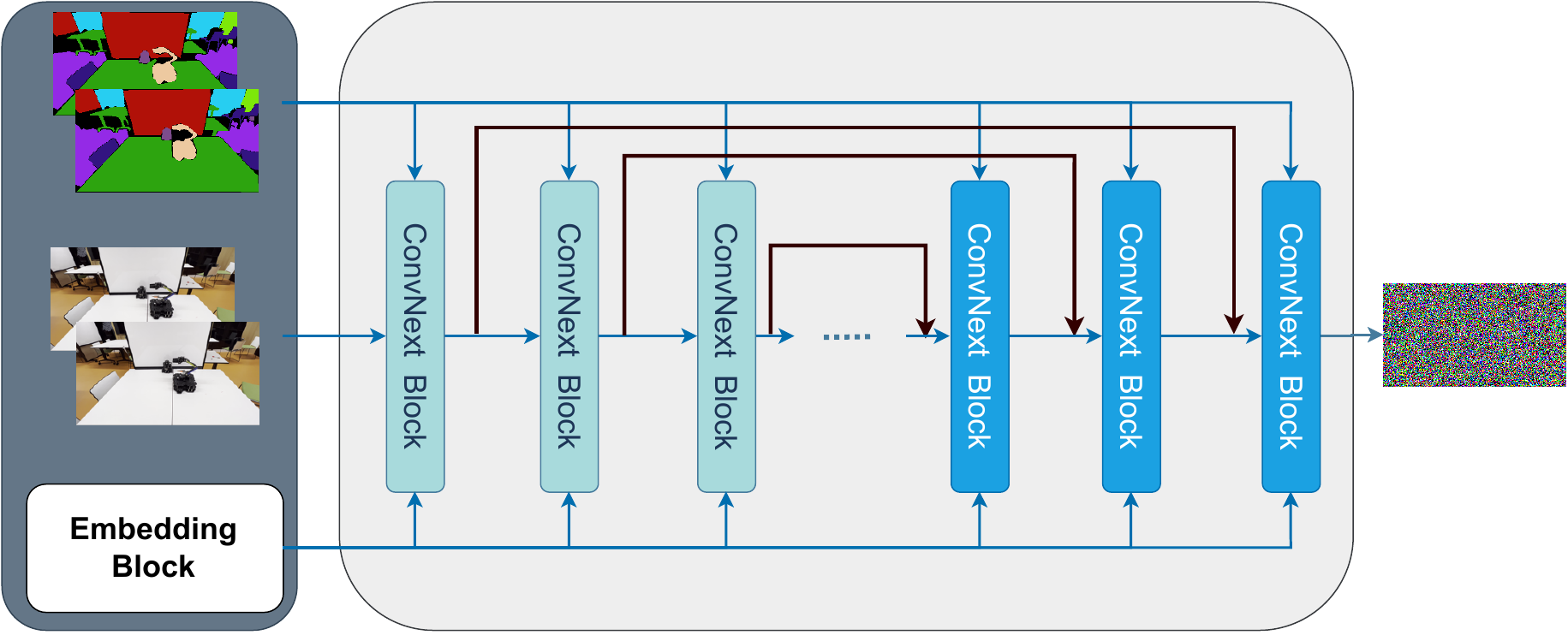}
    \caption{The overall architecture of the proposed diffusion model for shape and location retention. The black arrows indicate residual connections. It is worth noting we use images that depict masks of both the robot and the background. However, we also consider cases where only robot masks are used in this paper.}
    \label{fig:pipeline}
\end{figure*}

\subsection{Overall Architecture}
\begin{figure}[tb]
    \centering \includegraphics[width=0.75\linewidth]{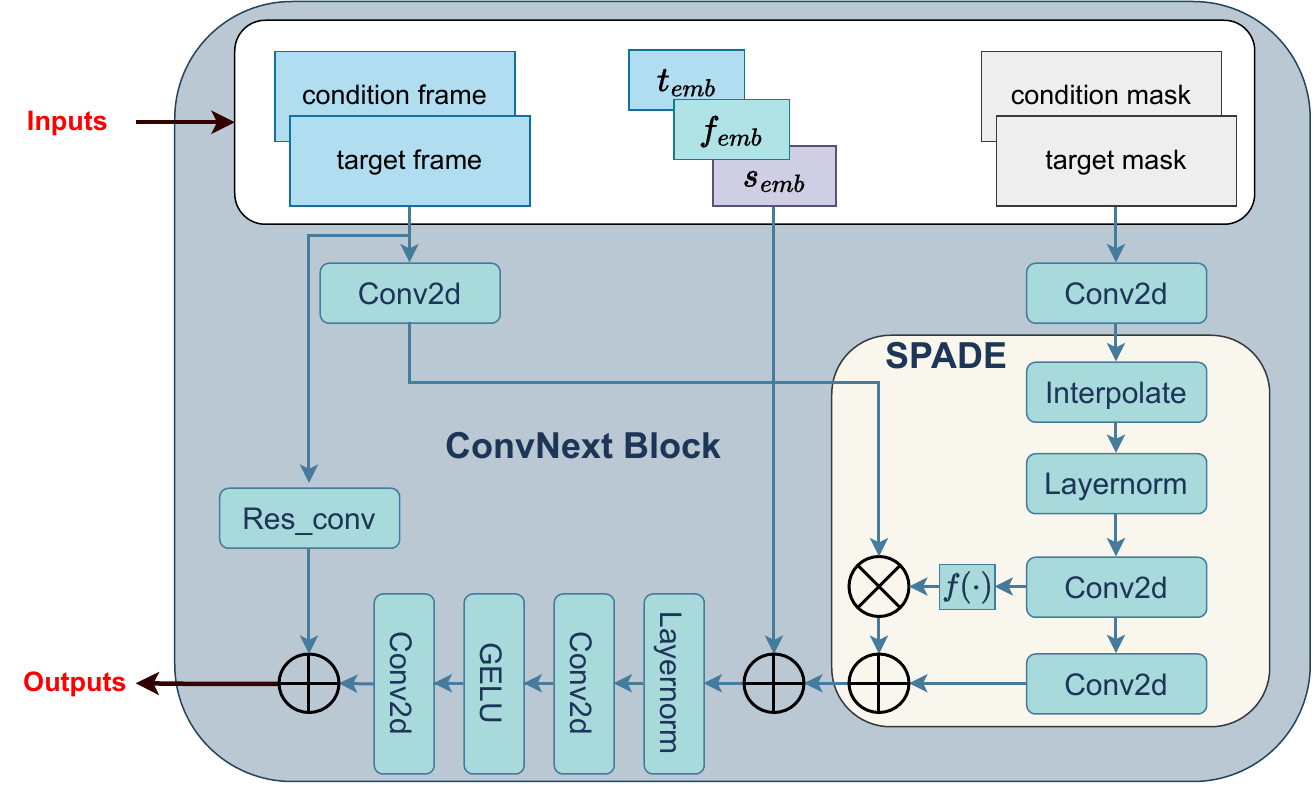}
    \caption{The improved ConvNext block with a SPADE module. The symbol $\otimes$ represents element-wise products and $\oplus$ indicates the sum of two tensors.}
    \label{fig:convnext}
\end{figure}

Figure~\ref{fig:pipeline} shows the overall architecture of the proposed shape and location retaining diffusion models, as well as the inputs and outputs of the model. We have adopted the ConvNext~\cite{liu2022convnet} comprised of standard ConvNet modules as the backbone network, which has been proven to be efficient while still facing challenges of containing distorted objects in generated frames~\cite{nikankin2022sinfusion}. To this end, we have introduced semantic mask regulation and robot pose embedding into the module, to improve shape and location retention of such models. The mask regulation and robot pose embedding modules are depicted in Figure~\ref{fig:convnext} and Figure~\ref{fig:embedding}, respectively. More details are given as follows.

\subsection{Inputs and Robot Pose Embedding}
\begin{figure}[tb]
    \centering \includegraphics[width=0.65\linewidth]{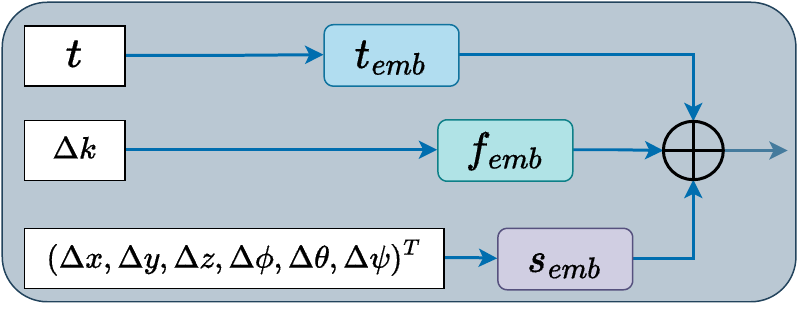}
    \caption{The embedding block in Fig. \ref{fig:pipeline}. We use $(\Delta x, \Delta y, \Delta z, \Delta \phi, \Delta \theta, \Delta \psi)^T$ to represent the robot pose difference between the condition frame and the current frame (frame to generate), the frame index difference is denoted as $\Delta k$, and the diffusion time step is denoted as $t$.}
    \label{fig:embedding}
\end{figure}

The inputs of the model include 1)~ A condition frame $\mathbf{x}_0^n$ sampled from a video comprising $N$ frames $\{\mathbf{x}_0^1,\mathbf{x}_0^2,\cdots,\mathbf{x}_0^N\}$, along with a noisy frame $\mathbf{x}_t^{n+\Delta k}$ where $t$ denotes the diffusion steps of $\mathbf{x}_0^n$, and $\Delta k$ represents the frame difference between $\mathbf{x}_0^n$ and $\mathbf{x}_0^{n+\Delta k}$. These frames are concatenated along the channel dimension as the first input. 2)~The diffusion time steps $t$ and frame index difference $\Delta k$ between the condition frame and the current frame are embedded following Equation (\ref{eq:sinuemb}).
\begin{equation}\label{eq:sinuemb}
\begin{aligned}
\gamma(p) = & \Big(\sin \left(2^0 \pi p\right), \cos \left(2^0 \pi p\right), \cdots, \\
& \sin \left(2^{L-1} \pi p\right), \cos \left(2^{L-1} \pi p\right)\Big),
\end{aligned}
\end{equation}
where $p$ represents either $t$ or $\Delta k$.  We have also embedded the robot pose difference vector $(\Delta x, \Delta y, \Delta z, \Delta \phi, \Delta \theta, \Delta \psi)^T:=\Delta \mathbf{P}$ into each ConvNext block, as shown in Figure~\ref{fig:embedding}. The motive behind this is to use robot pose information to guide model training, and when the model is trained, it will gain better shape and location retention performance in generated frames when conditioned on robot pose. In this paper, we use a linear embedding strategy for the pose difference vector embedding, i.e., $\Delta \mathbf{P}^\prime = \mathbf{A}\cdot \Delta \mathbf{P} + \mathbf{b}$. The motive behind this is the pose of the robot changes almost linearly as the time between two frames is short, e.g., 1 second, or $1/24$ seconds.

\subsection{Mask Regulation}
As mentioned earlier, using diffusion models to generate data with shape and location retention will benefit dangerous human-robot interaction by avoiding collecting data directly from real cases, which faces legal and ethical challenges. Semantic masks, abundant in shape and location information, have become easily accessible with advancements in object segmentation models like the Segment Anything model~\cite{kirillov2023segment}. Recognizing the potential benefits of leveraging semantic mask information, we propose incorporating it into ConvNext blocks to regulate intermediate outputs. Our approach introduces a new SPADE-based ConvNext block, outlined in Figure~\ref{fig:convnext}. Initially, frames and masks undergo separate processing through convolutional layers ($\text{conv2d}$), yielding outputs denoted as $\mathbf{x}$ and $\mathbf{m}$, respectively. Subsequently, the output $\mathbf{m}$ undergoes further processing using the proposed SPADE block to regulate $\mathbf{x}$. The SPADE block, as shown in Figure~\ref{fig:convnext}, is defined as follows.

\begin{equation}\label{eq:spade}
    \overline{\mathbf{x}}= \mathbf{x}\otimes f(\boldsymbol{\gamma}) \oplus \boldsymbol{\sigma},
\end{equation}
\noindent
where the symbol $\otimes$ indicates element-wise products, $f(\cdot)$ represents a mapping, $\overline{\mathbf{x}}$ is the output of the SPADE block, 
\begin{equation}\label{eq:layernorm}
    \overline{\mathbf{m}} = \text{Layernorm}\Big(\text{Interpolate}(\mathbf{m})\Big),
\end{equation}
\noindent
$\boldsymbol{\gamma}=\text{conv2d}(\overline{\mathbf{m}})$, and $\boldsymbol{\sigma}=\text{conv2d}(\boldsymbol{\gamma})$. We use the $\text{Layernorm}(\cdot)$ module to retain information from all mask channels and the nearest neighbor interpolation method is used for the $\text{Interpolate}(\cdot)$ module to ensure the size of masks matches that of the frames.

It is worth mentioning that the SPADE normalisation in Equation (\ref{eq:spade}) is different from~\cite{nikankin2022sinfusion} and \cite{park2019semantic} as we focus on using mask information to regulate intermediate outputs of ConvNext module such that shape and location information can be retained in video generation.

\subsection{Sampling}

 In the sampling phase, the model is presented with a singular frame extracted from the video to generate subsequent frames interactively. This process continues until the desired number of frames has been produced. During each iteration, the model utilises the provided frame and conditions such as pose information and semantic masks to inform the generation of the subsequent frames, ensuring a coherent and sequential flow of frames in the generated video.

\section{Experiments}
\subsection{Datasets}

Given the necessity of robot pose information for training the proposed models, we constructed our datasets accordingly. We employed ROS to control robots in diverse environments, capturing video footage at a frame rate of 24 frames per second (fps). Subsequently, we processed this footage to produce videos with a reduced frame rate of 1 fps, ensuring noticeable changes in the robot's pose. Our dataset comprises two types of robots: the Turtlebot Waffle Pi robot and an industrial collaborative robot, a.k.a. cobot. For the Turtlebot, we recorded videos in two laboratory environments: one with the robot and a simple background (Scene I) and the other with a more complex background (Scene II). Additionally, we recorded the translational and rotational velocities of the robot to calculate robot pose difference vectors. The frames of these videos were annotated to generate the necessary masks. We also created a third dataset (Scene III) featuring the cobot using a similar procedure to test the adaptability and robustness of our models. It is worth noting that our models focus on retaining the shape and location of objects of interest, such as robots, rather than super-resolution or high-resolution frame generation. Therefore, irrespective of the original frame sizes, we resized both the frames and masks to dimensions $256\times 144$ to optimise computational resources and accelerate training. This also facilitates fair comparisons with benchmark models. Our dataset is publicly accessible on \href{https://app.roboflow.com/turtlebot-h8awt}{Roboflow}.

\subsection{Models} 

In this paper, we explore two types of conditions: masks and robot pose information. To comprehensively compare and understand how these different conditions impact the shape and location retention performance of diffusion models, we investigate three models: 1) Ours-Mask-Pose, where both masks and pose information are utilised as conditions; 2) Ours-Mask, where only masks are employed as conditions; and 3) Ours-Pose, where only pose information is used as a condition. SinFusion is employed as the benchmark model for performance evaluation and comparison.

All three of our models utilise a backbone ConvNext consisting of 16 improved blocks, as depicted in Figure~\ref{fig:convnext}. To ensure fair comparison, the benchmark model also employs 16 blocks but lacks pose embedding and mask regulation. Additionally, our models feature several key distinctions: 1) When masks serve as conditions (in Ours-Mask-Pose and Ours-Mask models), they are subjected to regulation via the proposed SPADE module, as illustrated in Figure~\ref{fig:convnext}. 2) In instances where robot poses are employed as conditions (in Ours-Mask-Pose and Ours-Pose models), the difference in robot pose between two frames is embedded and integrated into the model, as depicted in Figure~\ref{fig:embedding}. Table~\ref{tab:robotPose} presents some example data of the robot pose used for embedding. Notably, these data are retrieved from ROS, simplifying the access to robot pose information.

\begin{table}[t]
    \centering
    \begin{tabular}{c|c|c|c|c|c}
        \hline
        $\Delta x (m)$ & $\Delta y (m)$ & $\Delta z (m)$ & $\Delta \phi (rad)$ & $\Delta \theta (rad)$ & $\Delta \psi (rad)$ \\
        \hline
        $\cdots$ & $\cdots$ & $\cdots$ & $\cdots$ & $\cdots$ & $\cdots$ \\
        -1 & 0 & 0 & 0 & 0 & 0.251 \\
        -0.5 & 0 & 0 & 0 & 0 & 0.252 \\
        -2 & 1 & 0 & 0 & 0 & 0.252 \\
        $\cdots$ & $\cdots$ & $\cdots$ & $\cdots$ & $\cdots$ & $\cdots$ \\
        \hline
    \end{tabular}
    \caption{Examples of robot pose data. It is noteworthy that as the robots move on flat floors, there are no translational changes along $z$ axis ($\Delta z = 0$),  and there are no rotational changes along $x$ ($\Delta \phi=0$) and $y$ ($\Delta \theta=0$) axes. We keep these columns to make the models general to robots work in different environments.}
    \label{tab:robotPose}
\end{table}

\begin{figure*}[t]
    \centering
    \includegraphics[width=0.75\linewidth]{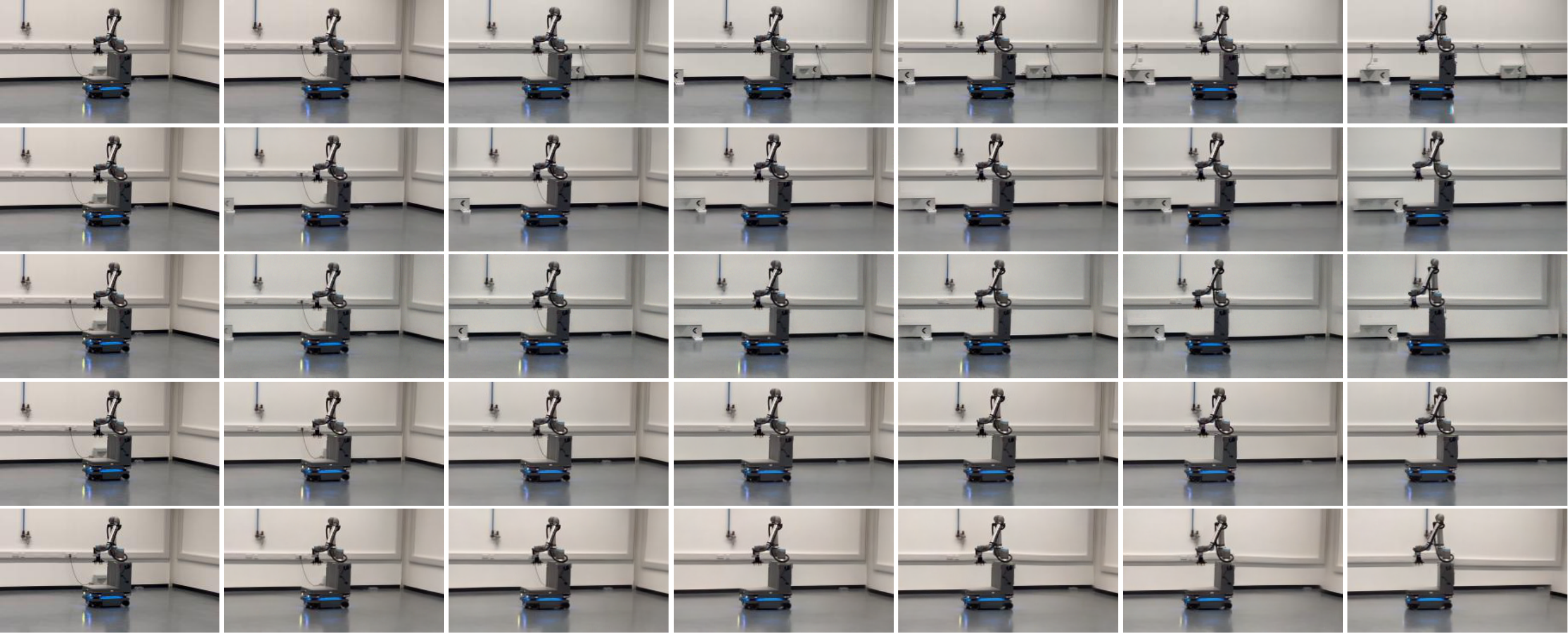}
    \caption[]{Results on Scene III, from top to bottom rows: 1) Original frames; 2) Ours-Mask-Pose; 3) Ours-Mask; 4) Ours-Poses; 5) SinFusion. It is noteworthy that only robot masks are used in models where masks are required for this set of results. While robot shape retention can be observed by comparing original and generated frames, location retention can be observed by comparing the robot location with the background in the upper right corner of the frames.}
    \label{fig:img_shown}
\end{figure*}

All models were trained on a single Nvidia A100 GPU. For our models, the loss function in Equation (\ref{eq:LexpandConditions}) is used for training, while SinFusion training employed Equation (\ref{eq:Lexpand}). Training durations varied among the models: the Ours-Mask-Pose model required approximately 6.2 hours, Ours-Mask took around 5.7 hours, and Ours-Pose took approximately 3.75 hours. In comparison, the benchmark model SinFusion required approximately 3.72 hours for training.

\begin{figure}[t]
    \centering
    \subfigure[]{\includegraphics[width=0.23\textwidth]{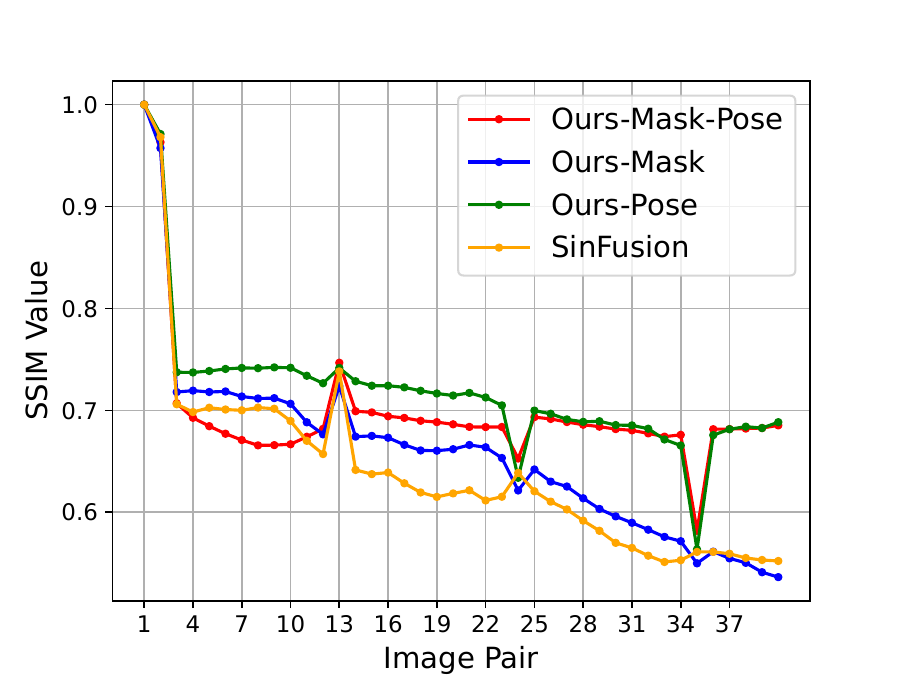}}
    \subfigure[]{\includegraphics[width=0.23\textwidth]{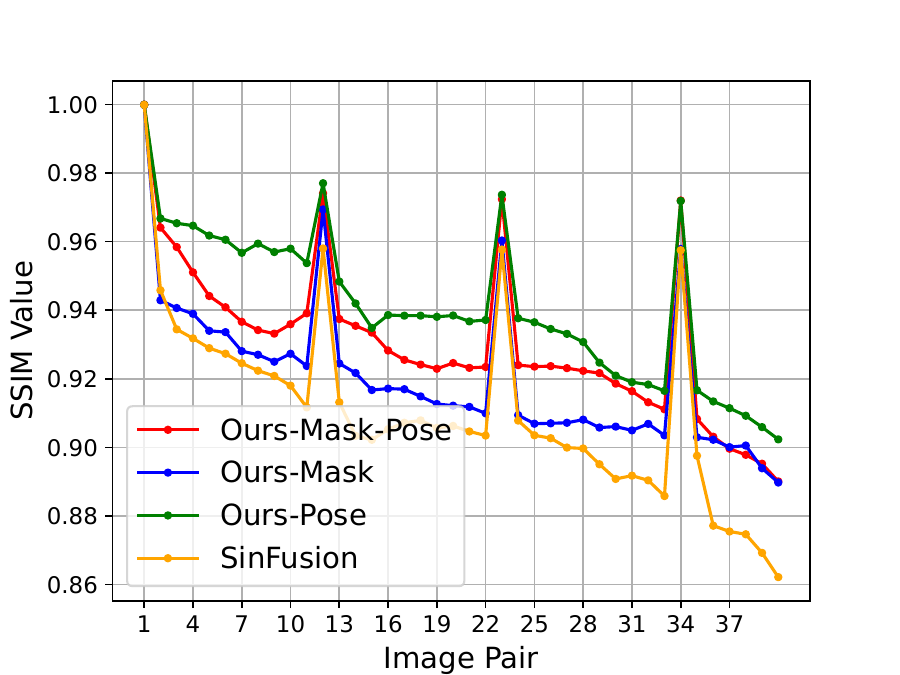}}
    \caption{Quality of generated frames against SSIM. (a) Scene I; (b) Scene III. Note that only robot masks are used.}
    \label{fig:ssimtworesults}
\end{figure}

\begin{figure*}[t]
    \centering
    \subfigure[Scene I: Hu Moments]{\includegraphics[width=0.23\textwidth]{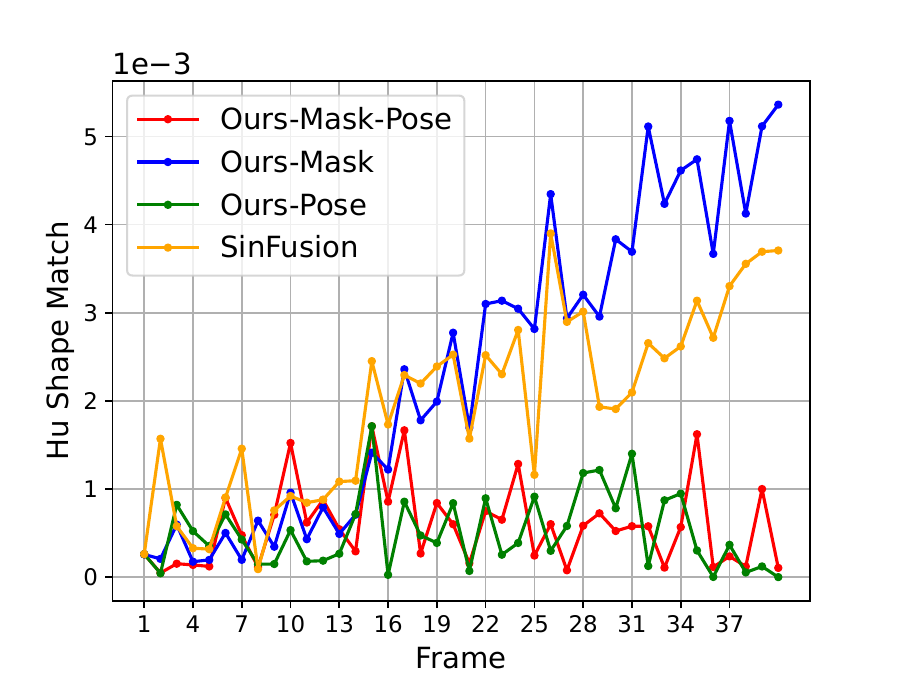}}
    \subfigure[Scene III: Hu Moments]{\includegraphics[width=0.23\textwidth]{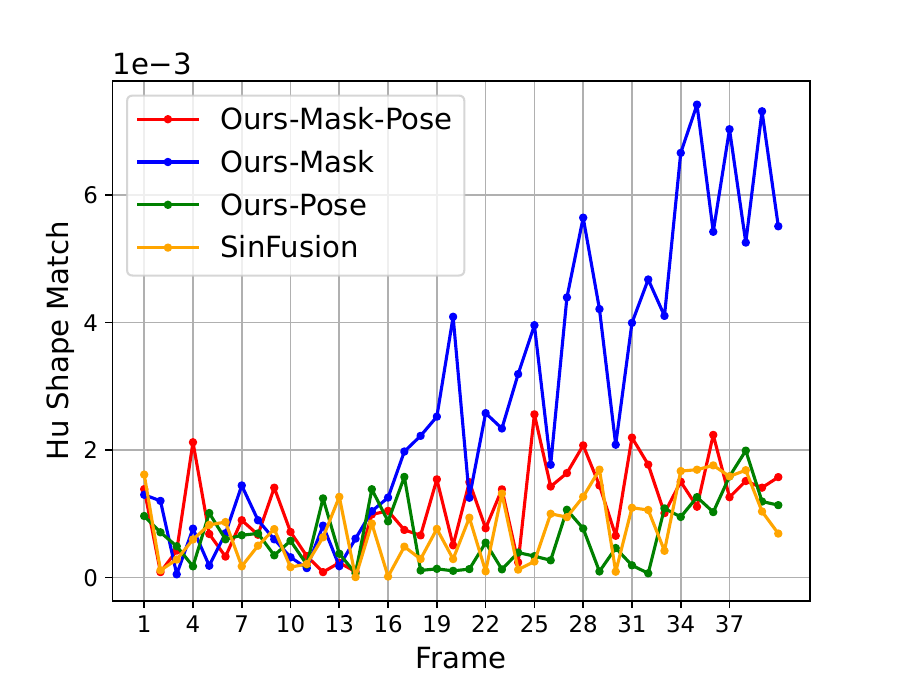}}
    \subfigure[Scene I: IoU]{\includegraphics[width=0.23\textwidth]{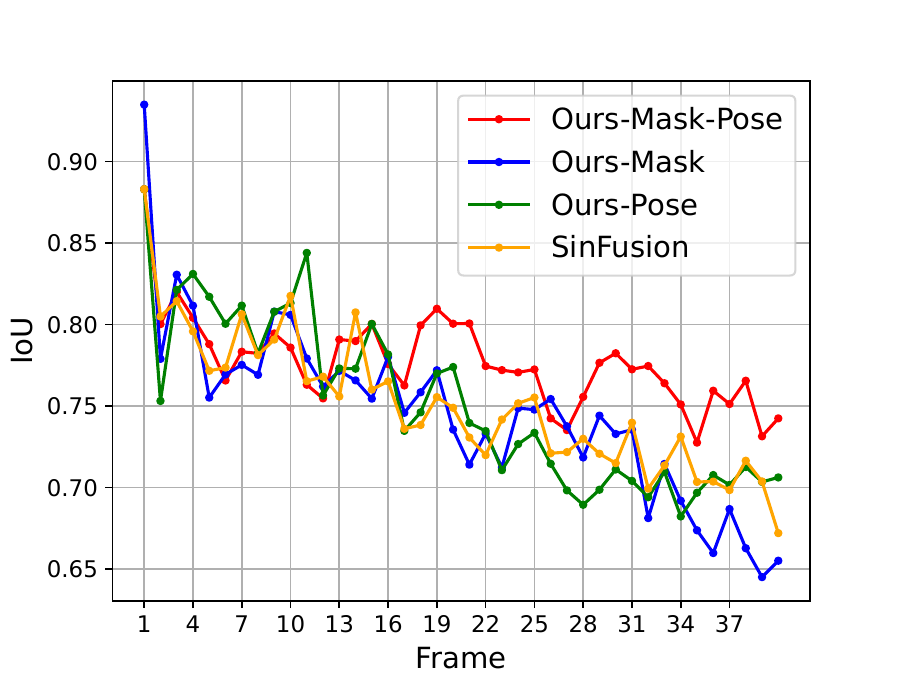}}
    \subfigure[Scene III: IoU]{\includegraphics[width=0.23\textwidth]{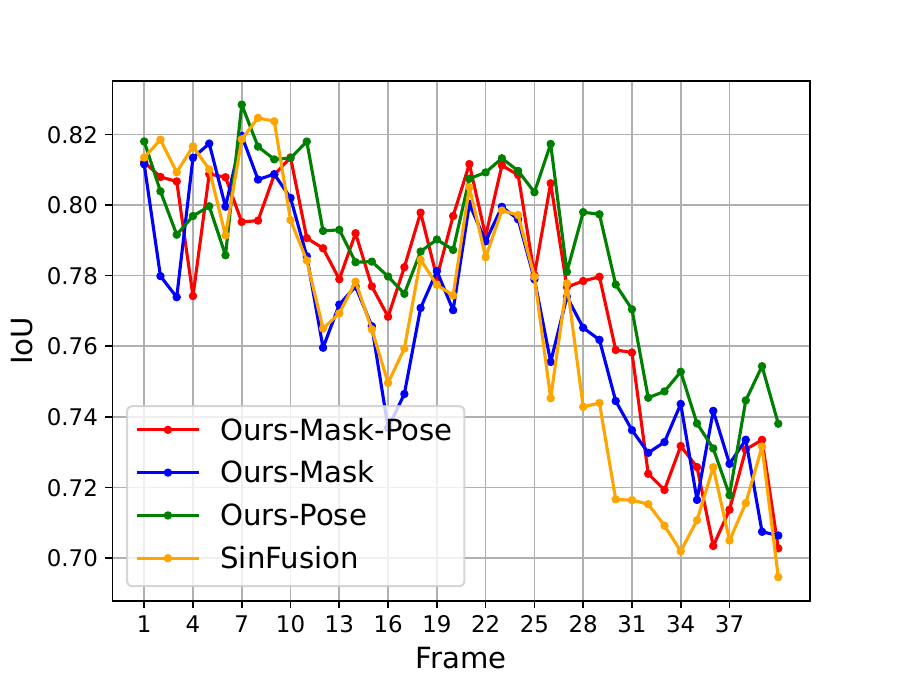}}
    \caption{Shape and location retention performance of different models. Only the robot masks are used where masks are used as conditions for frame generation.}
    \label{fig:huandiouresults}
\end{figure*}

\subsection{Evaluation Metrics} 
Three metrics are employed to assess model performance. The Structural Similarity Index (SSIM) is utilised to evaluate frame generation quality across different models. SSIM is preferred over PSNR (Peak Signal-to-Noise Ratio) for two main reasons: Firstly, SSIM measures image similarity in terms of structural information, luminance, and contrast, providing a more comprehensive assessment compared to PSNR, which solely quantifies reconstruction quality by comparing pixel values between original and generated frames. Secondly, as our focus is on retaining the shape and location of objects in generated frames, SSIM offers a more relevant comparison metric since shape and location information is assessed at a structural level rather than at the pixel level.

Shape-retention performance is evaluated by comparing the Hu moments of the $i$-th original frame $\mathbf{m}_\text{orig}^i$ with those of the $i$-th generated frame $\mathbf{m}_\text{gen}^i$. Hu moments are seven real-valued descriptors chosen for their ability to capture essential shape properties of an object of interest. These moments offer a concise representation of shape features, encompassing characteristics such as orientation, scale, and skewness~\cite{hu1962visual}. Equation (\ref{eq:hu}) is utilised to quantify the shape-retaining performance of diffusion models compared to the original video. The output of Equation (\ref{eq:hu}) indicates the dissimilarity between shapes in the generated frames and their corresponding original frames, with smaller values suggesting greater similarity. More information about Hu moments can be found in the \href{https://stummuac-my.sharepoint.com/:b:/g/personal/55141653_ad_mmu_ac_uk/EfeilbajY9ZAvnlgC9egMysBxTxxkrnkdymiOv1tR1taVA?e=IvBenj}{Supplemental Materials}.

\begin{equation}\label{eq:hu}
d^i = \sqrt{\sum_{j=1}^{7} \Big(\mathbf{M}_\text{orig}^i[j] - \mathbf{M}_\text{gen}^i[j]\Big)^2},
\end{equation}
\noindent
where $d^i$ is the similarity between shapes of interest in the $i$-th original and generated frames, and $\mathbf{M}_\text{orig}^i[j]$ and $\mathbf{M}_\text{gen}^i[j]$ represent the $j$-th Hu moments of the $i$-th original and generated frames, respectively.

The Intersection over Union (IoU) metric is utilised to assess the model's performance in retaining the robot's location. Rather than directly determining the precise location of the robot, we employ Equation (\ref{eq:iou}) to calculate the IoU between the masks of the robot in the $i$-th original frame $\mathbf{m}_\text{orig}^i$ and the mask of the robot in the $i$-th generated frame $\mathbf{m}_\text{gen}^i$. This computation serves as an indicator of how effectively the location is preserved in the generated videos.

\begin{equation}\label{eq:iou}
    \text{IoU}^i = \frac{\mathbf{m}_\text{orig}^i \bigcap \mathbf{m}_\text{gen}^i}{\mathbf{m}_\text{orig}^i \bigcup \mathbf{m}_\text{gen}^i}
\end{equation}

\subsection{Main Results} 

To delve deeply into the impact of masks and poses on the performance of shape and location retention, we have first considered the masks of the two types of robots exclusively. Two sets of experiments were conducted, one using the Scene I datasets and the other using the Scene III datasets.

The trained models from each dataset are employed to generate frames for evaluation. Figure~\ref{fig:img_shown} displays some of the generated results from Scene III, with additional results available in the \href{https://stummuac-my.sharepoint.com/personal/55141653_ad_mmu_ac_uk/_layouts/15/onedrive.aspx?id=%2Fpersonal%2F55141653%5Fad%5Fmmu%5Fac%5Fuk%2FDocuments%2FFaculty%2Ddoc%2FResearch%2FPeng%2DWang%2FIROS%5F2024%2FSupplemental%20materials%20to%20%20Robot%20Shape%20and%20Location%20Retention%20in%20Video%20Generation%20Using%20Diffusion%20Models%2Epdf&parent=%2Fpersonal%2F55141653%5Fad%5Fmmu%5Fac%5Fuk%2FDocuments%2FFaculty%2Ddoc%2FResearch%2FPeng%2DWang%2FIROS%5F2024&ga=1}{Supplemental Materials} due to space limit.

Quantitative evaluation results using the three metrics are computed: 1) shape retention based on Equation (\ref{eq:hu}); 2) location retention based on Equation (\ref{eq:iou}); and 3) overall quality of generated frames based on SSIM. The SSIM results are depicted in Figure~\ref{fig:ssimtworesults}, while the shape and location retention results are illustrated in Figure~\ref {fig:huandiouresults}. 

\begin{figure}[t]
    \centering
    \subfigure[]{\includegraphics[width=0.23\textwidth]{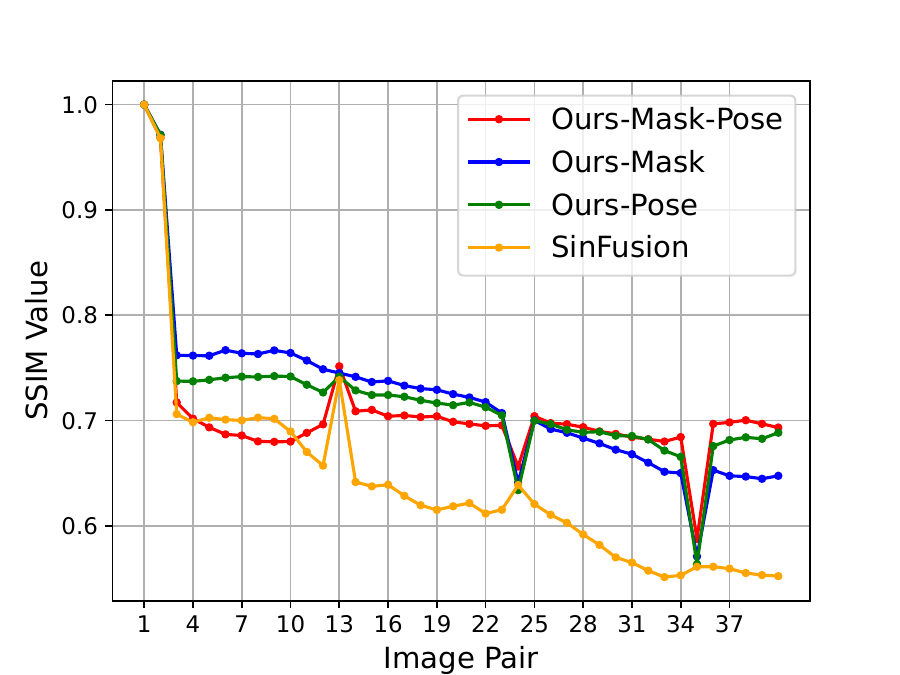}}
    \subfigure[]{\includegraphics[width=0.23\textwidth]{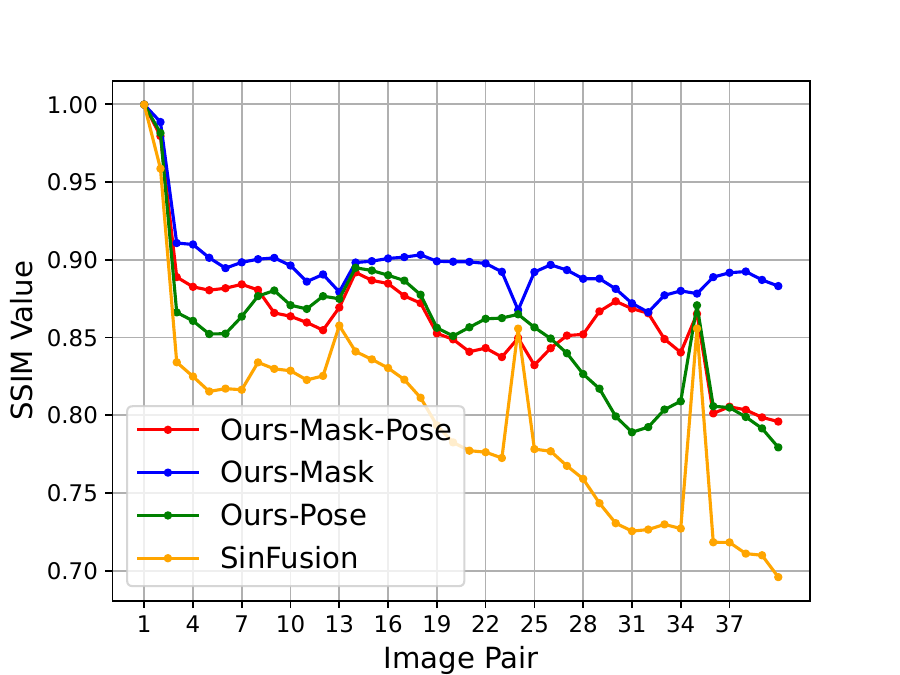}}
    \caption{Quality of generated frames against SSIM. (a) Scene II; (b) Scene III. Both robot and background masks are used.}
    \label{fig:ssimtwoallmaskresults}
\end{figure}

\begin{figure*}[b]
    \centering
    \subfigure[Scene II: Hu Moments]{\includegraphics[width=0.23\textwidth]{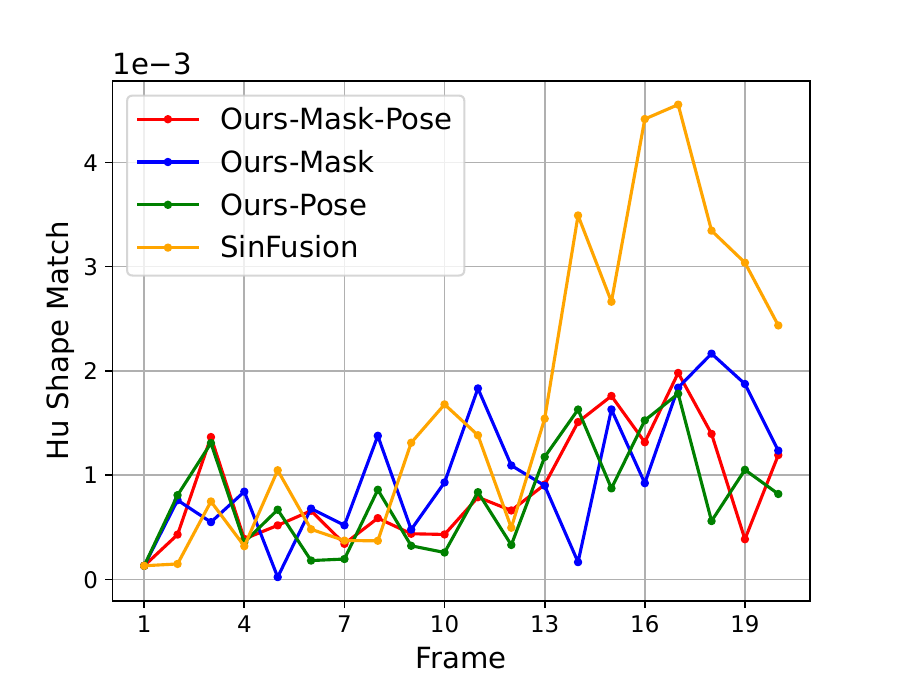}}
    \subfigure[Scene III: Hu Moments]{\includegraphics[width=0.23\textwidth]{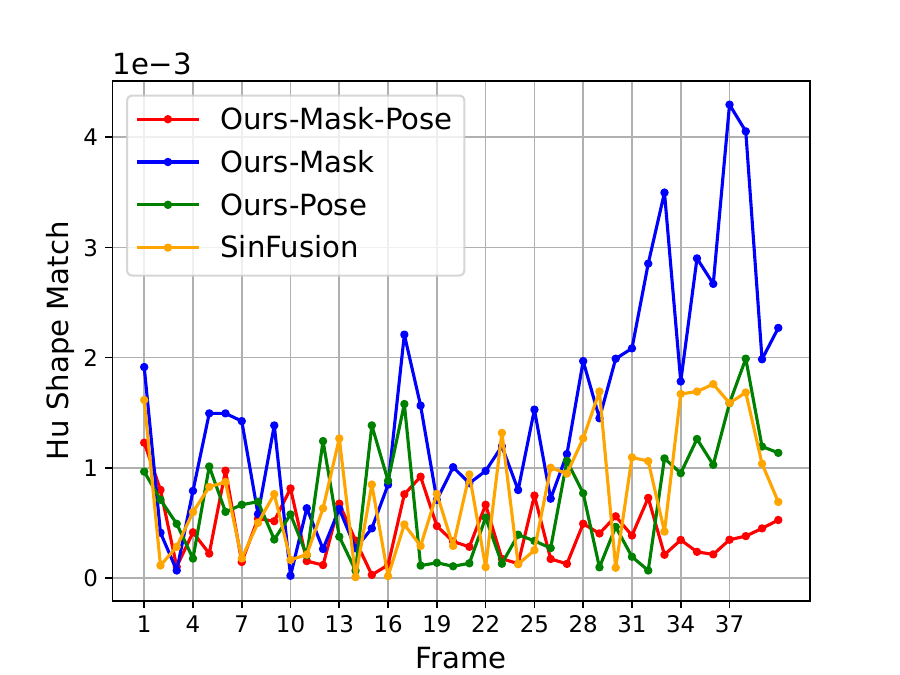}}
    \subfigure[Scene II: IoU]{\includegraphics[width=0.23\textwidth]{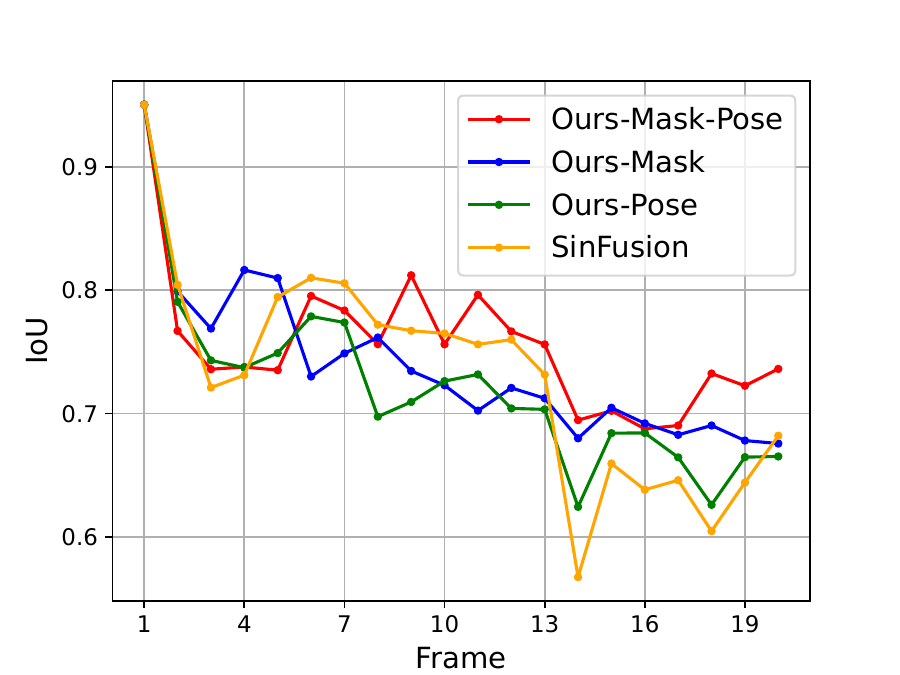}}
    \subfigure[Scene III: IoU]{\includegraphics[width=0.23\textwidth]{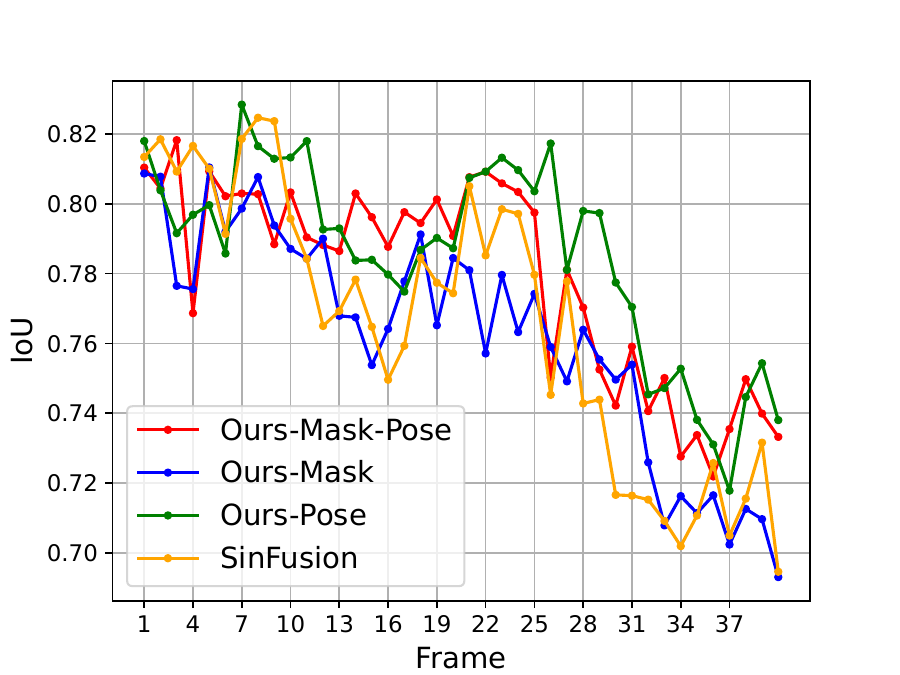}}
    \caption{Shape and location retention performance of different models. Both robot and background masks are used where masks are employed as conditions for frame generation.}
    \label{fig:huiouallsegsresults}
\end{figure*}
Regarding the overall quality of the generated frames, it can be observed from Figure~\ref{fig:ssimtworesults} that Ours-Pose achieves the best results, and Ours-Mask-Pose achieves comparable results, but both outperform the benchmark model. Regarding shape and location retention, it is evident from Figure~\ref{fig:huandiouresults} that Ours-Mask-Pose achieves either the best or the second-best results in both aspects. Ours-Pose achieves comparable results with Ours-Mask-Pose in shape retention. In terms of location retention, Ours-Mask-Pose performs comparably with Ours-Pose and outperforms other models in both Scene I and Scene III. In conclusion, incorporating sole pose information or the combination of pose information with masks improves the performance of diffusion models compared to the benchmark model across all three metrics. However, considering only mask results does not always improve the performance compared to the benchmark models, which we assume is due to the exclusive use of robot masks. Further experiments are conducted to investigate this phenomenon.

\subsection{Ablation Study}
\subsubsection{Considering Both Robot and Background Masks}
To further investigate the impact of masks on the generation results, additional experiments were conducted on Scene II and Scene III, using masks of both the robots and the backgrounds as conditions. The SSIM results are presented in Figure~\ref{fig:ssimtwoallmaskresults}, while the shape and location retention results are depicted in Figure~\ref{fig:huiouallsegsresults}. It is evident that by considering both robot and background masks, the quality of generated frames by Ours-Mask has improved in terms of SSIM. In Scene II, Ours-Mask achieves comparable results with Ours-Pose or Ours-Mask-Pose, and in Scene III, it either slightly outperforms Ours-Pose and Ours-Mask-Pose or achieves comparable results. Regarding shape and location retention, improvements are observed with Ours-Mask as well, as shown in Figure~\ref{fig:huiouallsegsresults}. However, Ours-Pose and Ours-Mask-Pose still outperform Ours-Mask in both shape and location retention in both scenes.

\begin{figure}[t]
    \centering
    \includegraphics[width=1.0\linewidth]{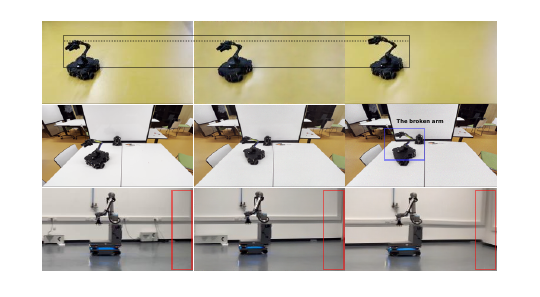}
    \caption{Evalution. \textit{Left:} the original frame; \textit{Middle:} Ours-Mask-Pose; \textit{Right}: SinFusion. \textit{Top row:} Ours-Mask-Pose retains the arm shape better; \textit{Middle row:} the robot arm was broken in the frame generated by SinFusion; \textit{Bottom row:} Ours-Mask-Pose retains the location of the robot better. }
    \label{fig:irosdetails}
\end{figure}

\subsubsection{The implication of Shape and Location Retention}
Some examples of shape and location retention of the robots are provided in Figure~\ref{fig:irosdetails}. In Scene I, Ours-Mask-Pose keeps the shape of the robot better compared to the benchmark model. In Scene II, similar results are observed and the robot arm is broken into two in the generated frame by the benchmark model. The location retention is shown in the results from Scene III, this can be recognised from the relative location of the robot and the wall highlighted.

Considering all experiments across the three scenes, it can be concluded that masks and pose information contribute to retaining the structural information of generated frames. In the meantime, it is important to highlight that models incorporating robot pose embedding only have consistently achieved comparable results in terms of location retention to those incorporating mask regulation, albeit with shorter training times. However, considering robot and/or background masks helps to improve the performance in shape retention and SSIM, but normally needs a longer model training time. Regardless, better performance has been achieved by the proposed models compared to the benchmark model.

\subsection{Discusions} 
Considering the performance of our models and the benchmark models against the metrics SSIM, IoU, and Hu shape similarity, this work has attempted to provide a solution to assess model performance in shape and location retention. Experimental results in different scenes of two types of robots show that taking robot pose information and mask (robot masks, or both robot and background masks) as conditions help to achieve significant improvements in all scenes against the metrics. We believe that shape and location retention in generated frames will benefit hazardous human-robot interaction detection in generating data for detection model training and beyond, which will be investigated in future works.

\section{CONCLUSIONS}

This paper introduces diffusion models that leverage robot pose and masks as conditional inputs for video generation. The objective is to produce video frames that maintain high structural fidelity, thereby enhancing the preservation of the shape and location information of objects within the generated frames. Through a series of experiments conducted across three distinct scenes involving various robots, we consistently observed improvements in generation quality as measured by SSIM, as well as in the retention of shape and location evaluated using Hu moments and IoU. These advancements hold promise for applications where accurate depiction of robot shape and location is crucial. For instance, our models can generate data to facilitate accurate dangerous human-interaction detection training, which will help mitigate potential risks associated with human-robot interactions.

\bibliographystyle{IEEEtran}
\bibliography{IROS2024DIFFUSION}

\end{document}